\documentclass[conference]{IEEEtran}
\IEEEoverridecommandlockouts
\usepackage{amsmath,amssymb,amsfonts}
\usepackage{algorithmic}
\usepackage{graphicx}
\usepackage{textcomp}
\usepackage[dvipsnames]{xcolor}
\usepackage{multirow}
\usepackage{multicol}
\usepackage{url}
\usepackage[
    backend=biber,
    style=science,
]{biblatex}
\providecommand{\keywords}[1]
{
  \small	
  \textbf{\textit{Keywords---}} #1
}

\begin{document}

\title{Text Mining Drug/Chemical-Protein Interactions using an Ensemble of BERT and T5 Based Models\\
}

\author{
Virginia Adams\textsuperscript{\textsection},
Hoo-Chang Shin\textsuperscript{\textsection},
Carol Anderson\textsuperscript{\textsection},
Bo Liu\textsuperscript{\textsection},
Anas Abidin\textsuperscript{\textsection},
\\
NVIDIA / Santa Clara, California, USA\\
\texttt{\{vadams;hshin;carola;boli;aabidin\}@nvidia.com}
}

\maketitle

\begingroup\renewcommand\thefootnote{\textsection}
\footnotetext{In reverse alphabetical order - authors contributed equally.}
\endgroup

\begin{abstract}
In Track-1 of the BioCreative VII Challenge participants are asked to identify interactions between drugs/chemicals and proteins. In-context named entity annotations for each drug/chemical and protein are provided and one of fourteen different interactions must be automatically predicted. For this relation extraction task, we attempt both a BERT-based sentence classification approach, and a more novel text-to-text approach using a T5 model. We find that larger BERT-based models perform better in general, with our BioMegatron-based model achieving the highest scores across all metrics, achieving 0.74 F1 score. Though our novel T5 text-to-text method did not perform as well as most of our BERT-based models, it outperformed those trained on similar data, showing promising results, achieving 0.65 F1 score. We believe a text-to-text approach to relation extraction has some competitive advantages and there is a lot of room for research advancement. \\
\end{abstract}


\section{Introduction}
The task of relation extraction, particularly for drug/chemical and protein, can be useful for many applications.
For instance, finding if a chemical is a down-regulator or up-regulator of a protein can be useful for drug discovery.
Christopoulou et al.~\cite{christopoulou2020adverse} show using relation extraction for finding adverse drug events.

With this motivation, the DrugProt shared task in BioCreative VII challenge/workshop is to find the relation between drug/chemical and proteins from biomedical literature in PubMed.
The task is to classify drug/chemical and protein relation into 13 possible classes, where the named entities of drug/chemical and proteins (candidates for the relations) are provided as annotated.

There was a previous related ChemProt task \cite{taboureau2010chemprot}, where the data format and task are similar.
DrugProt task has more data and the relation annotations are more granular.
Since the introduction of the ChemProt task, Lee et al.~\cite{lee2020biobert} and many others have shown the effectiveness of BERT~\cite{bert} models pre-trained in-domain PubMed data.
In addition, Beltagy et al.~\cite{beltagy2019scibert} and Gu et al.~\cite{gu2021domain} show the benefits of in-domain vocabulary set learned from PubMed literatures.

Furthermore, Shin et al.~\cite{biomegatron} show the additional benefit of larger model size with their BioMegatron models.
Here, the authors show that larger domain-specific language models outperform their smaller out-of-domain counterparts on a variety of biomedical natural language processing tasks. For our BioCreative VII Track-1 submission, we first repeat the BioMegatron study with the given Track-1 data, verifying the size and domain-specific hypothesis.

We then experiment with a novel text-to-text approach using T5~\cite{t5}. We convert the relation extraction task from sentence classification into a text-based question answering problem. We introduce novel ideas such as multi-step question answering and question balancing to improve performance.

Lastly, we use a model ensemble technique~\cite{ensemble} to boost the final performance of our submissions.

\section{Dataset, Pre-Processing and Software}
The DrugProt dataset \cite{drugprot} contains abstract, named entities of drug/chemical and protein pairs.
In the training and development set, the relation annotation for the pairs are also provided.
We conduct minimal pre-processing where we \textit{(i)} break the abstracts into sentence-level, and \textit{(ii)} sub-tokenize the words as in \cite{bert,t5}. 

We use Pandas data library \footnote{\url{https://pandas.pydata.org/}} for pre-processing, and PyTorch-based Megatron-LM~\cite{shoeybi2019megatron} and NeMo \footnote{https://github.com/NVIDIA/NeMo} codebase for further pre-processing, training, and testing.

\section{BERT-Based Models}
Our BERT-based models use the widely adopted relation extraction approach of annotating entities with special tokens and performing sentence classification. Formulated as multi-class classification, there are 14 total classes -- 13 relation classes and an additional no-relation class. We also try the recent advanced method of “matching the blanks” ~\cite{matching} using open-sourced code~\cite{matching-code}, but find the benefits of model size and a domain-specific pre-training corpus outweigh the benefit of a novel training scheme.

\begin{figure*}
\centering
\includegraphics[width=1\textwidth]{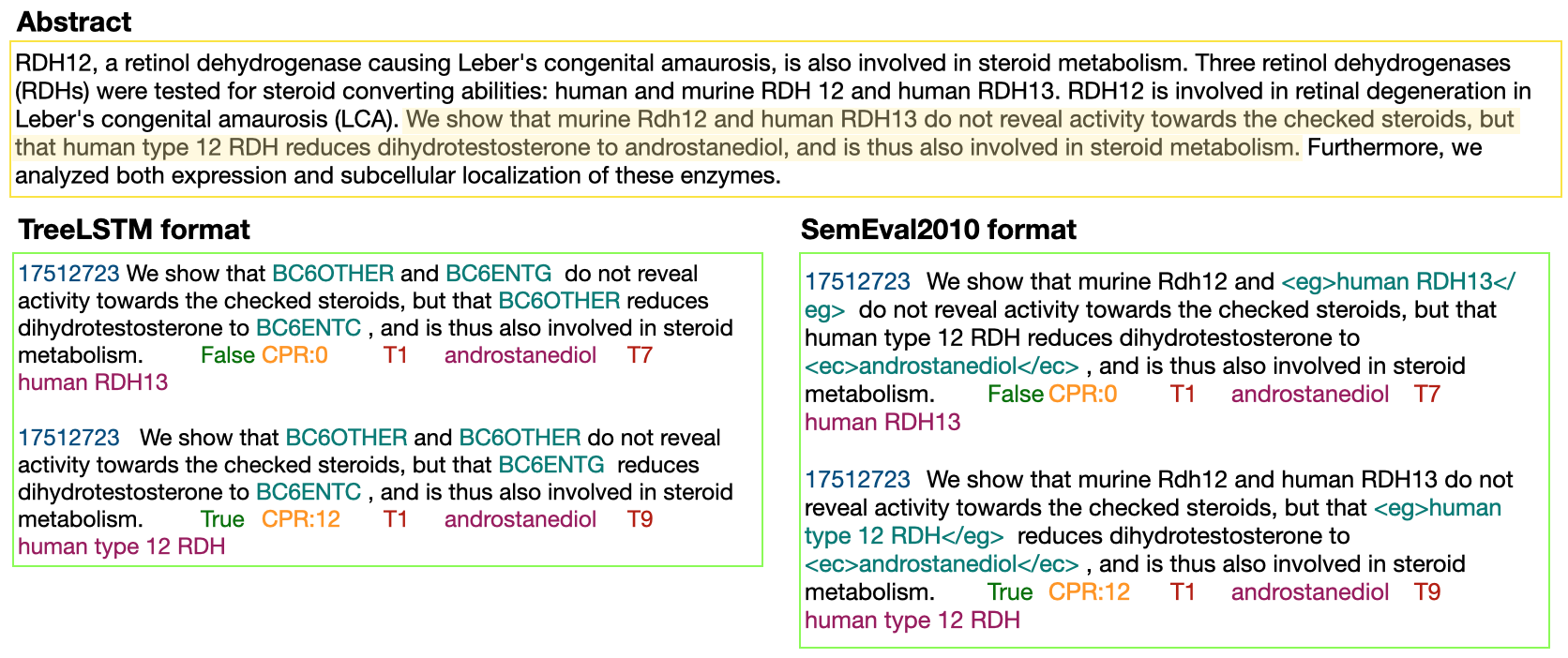}
\caption{An example converting the original text and dataset for multi-relation sentence classification into our BERT fine-tuning format. In the TreeLSTM \cite{treelstm} format, we mask the entities of interest with special chemical ({\color{teal}BC6ENTC}) and gene ({\color{teal}BC6ENTG}) tokens, while masking other chemicals and genes not under consideration with the special ``other'' token ({\color{teal}BC6OTHER}). {\color{OliveGreen}False} {\color{orange}CPR:0} indicates there is no relationship between the two entities under examination while {\color{OliveGreen}True} {\color{orange}CPR:12} signifies the true relationship's index is twelve, corresponding to "product-of". In these examples {\color{BrickRed}T1} indicates {\color{RedViolet}androstanediol}, the first entity to appear in the abstract, and {\color{BrickRed}T9} {\color{RedViolet}human type 12 RDH} is the ninth entity in the abstract. In the SemEval2010 \cite{semeval2010} format, we surround the entities of interest with special chemical ({\color{teal}\textless ec\textgreater human type 12 RDH\textless/ec\textgreater}) and gene ({\color{teal}\textless eg\textgreater androstanediol\textless/eg\textgreater}) tags, leaving all other chemicals and genes found in the sentence unannotated.}
\label{fig:annot_formats}
\end{figure*}

Examples of converting the original text and dataset for multi-relation sentence classification is shown in Figure~\ref{fig:annot_formats}. We break each abstract into sentences and annotate entity pairs of interest with special tokens to facilitate relation extraction as a sentence classification task. We experiment with two slightly different entity annotation schemes: \textit{(i)} masking the entities with special chemical and gene tokens \cite{treelstm}, and \textit{(ii)} surrounding the entities with chemical and gene tags \cite{semeval2010}.

We attach a fully-connected layer to the final BERT pooling layer for the classification.
We experiment with \{1, 2\} number-of-layers and \{128, 256, 384, 512\} number of hidden-units for the fully-connected layer.
Other hyper-parameters we experiment are: max-sequence-length of \{128, 256, 512\}, dropout-rate of \{0.1, 0.5, 0.9\}, learning-rate of \{5e-6, 9e-6, 1e-5, 5e-5\}.
We use Adam optimizer~\cite{kingma2014adam} with cross-entropy loss and train for 5 to 40 epochs.
The BERT-cased vocabulary \cite{bert} is used due to the time limitation to conduct enough experiments by the submission deadline.

Development set BERT-based evaluation results are shown in Table~\ref{tab:bert-t5-based-models-perf}. Because the methods, vocabulary sets, and hyper-parameters are not consistent, it is not a completely controlled experiment.
Nonetheless, we can observe the general trend of \textit{(i)} domain-specific models (BioBERT, BioMegatron) and \textit{(ii)} larger model size contributing beneficially to improved performance.

\section{T5-Based Models}

\begin{figure*}
\centering
\includegraphics[width=1\textwidth]{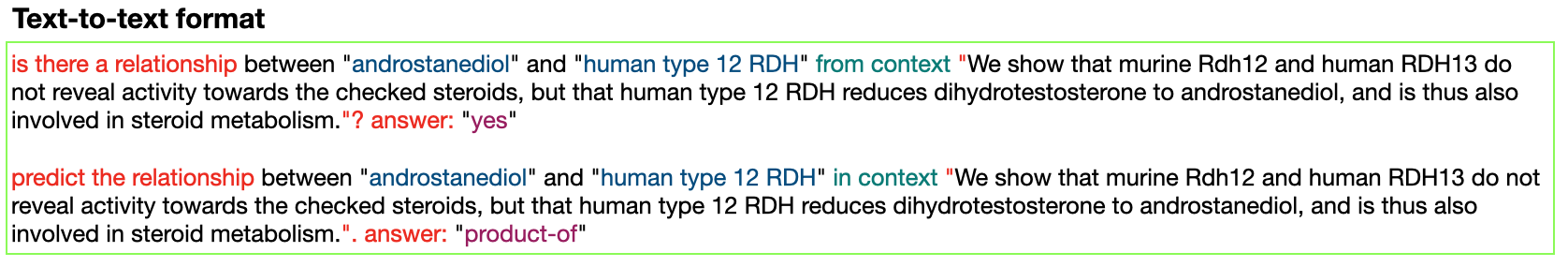}
\caption{An example of converting the original text and dataset into the text-to-text T5-base fine-tuning and evaluation format. The abstract is split into sentences. Each sentence is transformed into a sequence of question and answer pairs: \textit{first}, asking if a relationship between two specific entities is present in the sentence, and \textit{second}, prompting the model to predict the relationship if it indicated one exists.}
\label{fig:annot_formats-t5}
\end{figure*}

While BERT-based models have been a standard for some time now, more recent ``text-to-text'' or ``prompting'' based methods \cite{t5, prompting} have several advantages over transformer encoder language models with additional task specific architectures. One of the most notable advantages is the ability to perform multiple tasks without needing distinct specialized layers for each task. Through this formulation, every task can share the same cross-entropy minimization training objective and the inductive bias learned about biomedical entities for one task can possibly be transferred to another.  

A further advantage is the potential for ``zero-shot'' or ``few-shot'' learning \cite{prompting}. When tasks are expressed as text-based question answering problems, with sufficient model capacity, text-to-text generative models can produce impressive results on multiple tasks with little or no task-specific fine-tuning \cite{gpt3}. Though we fine-tune on the challenge data training set for this submission, we view this as the necessary first step in working towards our end goal of few-shot or zero-shot relation extraction.

\begin{table*}[htbp]
\caption{\label{tab:bert-t5-based-models-perf} Precision, Recall, and F-1 scores of T5 and BERT models on the development set.}
\centering
\resizebox{2\columnwidth}{!}{
\begin{tabular}{ccclcccc}
\hline
  \textbf{Model} & \textbf{Method} & \textbf{\#Parameters} & \textbf{Vocabulary} & \textbf{Prec}&\textbf{Rec}&\textbf{F-1} \\ \hline
  BERT-base    & \cite{matching} & 110m & BERT-uncased & 0.68 & 0.59 & 0.63 \\
  BioBERT-base & \cite{matching} & 110m & BERT-uncased & 0.71 & 0.67 & 0.69 \\
  BERT-base    & ---             & 110m & BERT-cased   & 0.77 & 0.58 & 0.66 \\
  BERT-large   & ---             & 345m & BERT-cased   & 0.74 & 0.70 & 0.72 \\
  \textbf{BioMegatron}  & ---  & \textbf{345m} & \textbf{BERT-cased} & \textbf{0.76} & \textbf{0.71} & \textbf{0.74} \\
\hline
  T5-base & over sampling positive & 345m & BERT-uncased & 0.36 & 0.78 & 0.49 \\
  T5-base & balancing negative/positive & 345m & BERT-uncased & 0.54 & 0.71 & 0.61 \\
  \textbf{T5-base} & \textbf{over sampling negative} & \textbf{345m} & \textbf{BERT-uncased} 
  &\textbf{0.67} & \textbf{0.63} & \textbf{0.65} \\
\hline
\end{tabular}}
\end{table*}

Many natural language processing tasks have been successfully reformulated as text-to-text tasks, such as text classification, natural language inference, summarization, and reading comprehension. To our knowledge there are no published studies to date that use a text-to-text approach for relation extraction, although a prompting-based approach using masked language modeling has been explored by Wei et al. \cite{knowprompt}.

We use T5 \cite{t5} for our text-to-text approach. Figure~\ref{fig:annot_formats-t5} shows an example of data conversion from the given biomedical abstract and entity annotations into the T5 prompting input and output. The abstract is split into sentences, and each sentence is turned into a sequence of natural language questions and answers. We first ask the model to identify if a specific relation is present in the sentence. If there is, we ask it to predict the relation. We investigated different prompt formats and empirically found this setup to yield the highest scores. We fine-tuned an off-the-shelf T5-base model that was pre-trained on general domain text via the methodology described by Raffel et al. \cite{t5}.

Evaluation results using T5 on the development set are shown in Table~\ref{tab:bert-t5-based-models-perf}. A noticeable improvement is achieved by balancing the positive and negative examples of sentences with and without relations and then over sampling the number of negative examples in the training set. Our best T5 model out performed our BERT-Base with BERT-uncased vocabulary model and performed within 0.01 F1 score of our BERT-Base with cased vocabulary model. BERT-Large models and BERT-Base models pre-trained on biomedical domain data out-score fine-tuned T5, but perhaps with in-domain pre-training and larger model capacity, T5 could further improve. 

Recent studies \cite{gpt3, flan} show that model size needs to be sufficiently large, such as having 5 billion parameters, in order to achieve good zero-/few-shot performance. Since our T5 models were relatively small (345 million parameters), we will definitely need to increase model size for few or zero shot relation extraction to be feasible. Nevertheless, our results are encouraging.

\section{Model Ensemble and Final Test-Set Scores}

Model ensembling \cite{ensemble} is a widely used technique to increase the final performance of machine learning models by combining multiple models' predictions, often averaging them.

For our final submission, we use an ensemble of different models and attain noticeable improvement in evaluation scores on both the development and test sets. We ensemble the models by taking a weighted average of each model's predicted probability vector then selecting the argmax from this averaged vector as our final prediction. This approach can work across a diverse set of models, even between our BERT and text-to-text models. In fact, ensembling diverse models is desirable because each single model's unique prediction errors can be overcome by generally low probability scores from the other models in the ensemble, masking individual model mistakes. 

We make four submissions in total. Ordered as in Table \ref{tab:final-perf}, our first submission is an ensemble of our fine-tuned BioBERT-Base and best T5 models. Our second submission is from our best T5 model alone. Third, we submit single model predictions from fine-tuned BioMegatron. Finally, our fourth and best submission as an ensemble of fine-tuned BERT-Base, BERT-Large, and BioMegatron.

\begin{table}[htbp]
\caption{\label{tab:final-perf}Final precision, Recall, and F-1 scores on the test set.}
\centering
\resizebox{1\columnwidth}{!}{
\begin{tabular}{lccc}
\hline
  \textbf{Model(s)} & \textbf{Prec}&\textbf{Rec}&\textbf{F-1} \\ \hline
  {[BioBERT, T5]}-ensemble & 0.71 & 0.67 & 0.69 \\
  T5                     & 0.64 & 0.58 & 0.61 \\
  BioMegatron            & 0.74 & 0.72 & 0.73 \\
  \textbf{{[BERT-base\&large, BioMegatron]}-ensemble} & \textbf{0.77} & \textbf{0.73} & \textbf{0.75} \\
\hline
\end{tabular}}
\end{table}

Table~\ref{tab:final-perf} shows the final official evaluation scores on the test set.
The best scores are achieved using an ensemble of multiple BERT-base\&-large, and BioMegatron models. Table~\ref{tab:final-perf-granular} shows the final official evaluation scores on the test set at a more granular level for each relation type. Some relations with only a few samples in the training data are generally difficult to classify correctly.

\begin{table}[htbp]
\caption{\label{tab:final-perf-granular}Granular scores for each relation type for two of our best-performing ensemble models.}
\centering
\resizebox{1\columnwidth}{!}{
\begin{tabular}{l|ccc|ccc}
\hline
                        & \multicolumn{3}{|c|}{{[BioBERT, T5]}} & \multicolumn{3}{c}{{[BERT, BioMegatron]}} \\
 \textbf{Relation-Type} & \textbf{Prec}&\textbf{Rec}&\textbf{F-1} & \textbf{Prec}&\textbf{Rec}&\textbf{F-1} \\
 \hline
 ACTIVATOR &	0.75	&0.64	&0.69&0.77&	0.76&	0.77\\
AGONIST	& 0.81	&0.73	&0.77&0.76&	0.722&	0.74\\
AGONIST-INHIBITOR &	1.0	&0.33	&0.5 & 0.0 & 0.0&	0.0\\
ANTAGONIST&	0.84	&0.87	&0.85&0.85&	0.92&	0.88\\
DIRECT-REGULATOR&	0.68	&0.56	&0.62&0.73&	0.65&	0.69\\
INDIRECT-DOWNREGULATOR&	0.62	&0.77&	0.69&0.74&	0.76&	0.75\\
INDIRECT-UPREGULATOR&	0.71	&0.67&	0.69&0.76&	0.78&	0.77\\
INHIBITOR	&0.75	&0.77	&0.76&0.84&	0.84&	0.84\\
PART-OF	&0.64	&0.66	&0.65&0.72&	0.59&	0.65\\
PRODUCT-OF&	0.64	&0.59	&0.61&0.70&	0.56&	0.62\\
SUBSTRATE&	0.63	&0.46&	0.53&0.68&	0.53&	0.59\\
SUBSTRATE\_PRODUCT-OF&	0.0	&0.0&	0.0&0.0&	0.0&	0.0\\
AGONIST-ACTIVATOR	&0.0	&0.0&	0.0&0.0&	0.0&	0.0\\
\hline
\end{tabular}}
\end{table}

\section{Large-scale Sub-Track}

For the large-scale track we did not perform additional model development due to time constraints.
We only remove pipelines unnecessary for inference from our smallest BERT-base model. This mostly includes convenience pipelines in PyTorch-Lightning. We then run inference on four GPUs, dividing the dataset into four sub-parts. Our resulting BERT-base model reports lower evaluation scores on the development set. It is possible we lost some performance due to lack of attention in model-stripping. Nonetheless, we finished inference on the large-scale sub-track test data.
The overall- and granular- official evaluation scores for the large-scale sub-track are shown in Table~\ref{tab:final-perf-large} and Table~\ref{tab:final-perf-granular-large}.

\begin{table}[htbp]
\caption{\label{tab:final-perf-large}Final precision, Recall, and F-1 scores on the large-scale sub-track.}
\centering
\resizebox{.5\columnwidth}{!}{
\begin{tabular}{lccc}
\hline
 \textbf{Model(s)} & \textbf{Prec}&\textbf{Rec}&\textbf{F-1} \\ \hline
 BERT-base & 0.73 & 0.33 & 0.46 \\
\hline
\end{tabular}}
\end{table}

\begin{table}[htbp]
\caption{\label{tab:final-perf-granular-large}Granular scores for each relation type on the large-scale sub-track.}
\centering
\resizebox{.8\columnwidth}{!}{
\begin{tabular}{l|ccc}
\hline
 \textbf{Relation-Type} & \textbf{Prec}&\textbf{Rec}&\textbf{F-1} \\
 \hline
ACTIVATOR               & 0.69  &  0.35  &  0.46 \\
AGONIST                 & 0.74  &  0.40   &  0.52 \\
AGONIST-INHIBITOR       & 0.0       &  0.0       &  0.0 \\
ANTAGONIST              & 0.75  &  0.48  &  0.58 \\
DIRECT-REGULATOR        & 0.66   &  0.19  &  0.30 \\
INDIRECT-DOWNREGULATOR  & 0.71  &  0.35  &  0.47 \\
INDIRECT-UPREGULATOR    & 0.70  &  0.43  &  0.53 \\
INHIBITOR               & 0.81  &  0.45  &  0.58 \\
PART-OF                 & 0.60  &  0.15  &  0.25 \\
PRODUCT-OF              & 0.62  &  0.23  &  0.33 \\
SUBSTRATE               & 0.69  &  0.17  &  0.27 \\
SUBSTRATE\_PRODUCT-OF    & 0.0       &  0.0       &  0.0 \\
AGONIST-ACTIVATOR       & 0.0       &  0.0       &  0.0 \\
\hline
\end{tabular}}
\end{table}

\section{Discussion}

Our results confirm previous findings that larger models tend to perform better than smaller ones, and models trained on domain-specific text tend to perform better than those trained on general domain data. For our BERT-based models, performance could potentially be improved beyond what we reported here by using even larger BioMegatron models, which we did not have time to complete and submit. For our T5 models, larger model size and pretraining on in-domain text would likely improve performance. We also confirm that model ensembling gives an additional performance boost, even when model architectures are different (BERT- and T5- based).

Although our text-to-text based methods did not perform as well as the largest BERT models we trained, their performance was similar to or better than BERT base models pretrained on general domain text. These results indicate that relation extraction can be successfully framed as a text-to-text task, while also highlighting some challenging aspects of the approach. In particular, we find that careful attention should be paid to class balancing during fine-tuning and to the design of prompts used for inference.

For the large-scale sub-track, we could only use our smallest BERT-base model.
Further improvement could be seen by applying advanced model optimization techniques such as quantization and pruning, allowing use of our larger models for the large-scale inference task.

\printbibliography

@article{bert,
  title={Bert: Pre-training of deep bidirectional transformers for language understanding},
  author={Devlin, Jacob and Chang, Ming-Wei and Lee, Kenton and Toutanova, Kristina},
  journal={arXiv preprint arXiv:1810.04805},
  year={2018}
}

@article{biomegatron,
  title={BioMegatron: Larger biomedical domain language model},
  author={Shin, Hoo-Chang and Zhang, Yang and Bakhturina, Evelina and Puri, Raul and Patwary, Mostofa and Shoeybi, Mohammad and Mani, Raghav},
  journal={arXiv preprint arXiv:2010.06060},
  year={2020}
}

@article{t5,
  title={Exploring the limits of transfer learning with a unified text-to-text transformer},
  author={Raffel, Colin and Shazeer, Noam and Roberts, Adam and Lee, Katherine and Narang, Sharan and Matena, Michael and Zhou, Yanqi and Li, Wei and Liu, Peter J},
  journal={arXiv preprint arXiv:1910.10683},
  year={2019}
}

@article{ensemble,
  title={Distilling the knowledge in a neural network},
  author={Hinton, Geoffrey and Vinyals, Oriol and Dean, Jeff},
  journal={arXiv preprint arXiv:1503.02531},
  year={2015}
}

@article{matching,
  title={Matching the blanks: Distributional similarity for relation learning},
  author={Soares, Livio Baldini and FitzGerald, Nicholas and Ling, Jeffrey and Kwiatkowski, Tom},
  journal={arXiv preprint arXiv:1906.03158},
  year={2019}
}

@misc{matching-code,
  howpublished = "\url{https://github.com/plkmo/BERT-Relation-Extraction}",
  note = "[Online; accessed 1-Sep-2021]"
}

@article{treelstm,
  title={Chemical--gene relation extraction using recursive neural network},
  author={Lim, Sangrak and Kang, Jaewoo},
  journal={Database},
  volume={2018},
  year={2018},
  publisher={Oxford Academic}
}

@article{semeval2010,
  title={Semeval-2010 task 8: Multi-way classification of semantic relations between pairs of nominals},
  author={Hendrickx, Iris and Kim, Su Nam and Kozareva, Zornitsa and Nakov, Preslav and S{\'e}aghdha, Diarmuid O and Pad{\'o}, Sebastian and Pennacchiotti, Marco and Romano, Lorenza and Szpakowicz, Stan},
  journal={arXiv preprint arXiv:1911.10422},
  year={2019}
}

@article{prompting,
  title={Pre-train, prompt, and predict: A systematic survey of prompting methods in natural language processing},
  author={Liu, Pengfei and Yuan, Weizhe and Fu, Jinlan and Jiang, Zhengbao and Hayashi, Hiroaki and Neubig, Graham},
  journal={arXiv preprint arXiv:2107.13586},
  year={2021}
}

@article{gpt3,
  title={Language models are few-shot learners},
  author={Brown, Tom B and Mann, Benjamin and Ryder, Nick and Subbiah, Melanie and Kaplan, Jared and Dhariwal, Prafulla and Neelakantan, Arvind and Shyam, Pranav and Sastry, Girish and Askell, Amanda and others},
  journal={arXiv preprint arXiv:2005.14165},
  year={2020}
}

@article{flan,
  title={Finetuned language models are zero-shot learners},
  author={Wei, Jason and Bosma, Maarten and Zhao, Vincent Y and Guu, Kelvin and Yu, Adams Wei and Lester, Brian and Du, Nan and Dai, Andrew M and Le, Quoc V},
  journal={arXiv preprint arXiv:2109.01652},
  year={2021}
}

@misc{knowprompt,
      title={KnowPrompt: Knowledge-aware Prompt-tuning with Synergistic Optimization for Relation Extraction}, 
      author={Xiang Chen and Ningyu Zhang and Xin Xie and Shumin Deng and Yunzhi Yao and Chuanqi Tan and Fei Huang and Luo Si and Huajun Chen},
      year={2021},
      eprint={2104.07650},
      archivePrefix={arXiv},
      primaryClass={cs.CL}
}

@misc{drugprot,
  title={Overview of DrugProt BioCreative VII track: quality evaluation and large scale text mining of drug-gene/protein relations},
  author={Miranda, Antonio and Mehryary, Farrokh and Luoma, Jouni and Pyysalo, Sampo and Valencia, Alfonso and Krallinger, Martin},
  journal={Proceedings of the seventh BioCreative challenge evaluation workshop},
  year={2021}
}

@article{christopoulou2020adverse,
  title={Adverse drug events and medication relation extraction in electronic health records with ensemble deep learning methods},
  author={Christopoulou, Fenia and Tran, Thy Thy and Sahu, Sunil Kumar and Miwa, Makoto and Ananiadou, Sophia},
  journal={Journal of the American Medical Informatics Association},
  volume={27},
  number={1},
  pages={39--46},
  year={2020},
  publisher={Oxford University Press}
}

@article{taboureau2010chemprot,
  title={ChemProt: a disease chemical biology database},
  author={Taboureau, Olivier and Nielsen, Sonny Kim and Audouze, Karine and Weinhold, Nils and Edsg{\"a}rd, Daniel and Roque, Francisco S and Kouskoumvekaki, Irene and Bora, Alina and Curpan, Ramona and Jensen, Thomas Sk{\o}t and others},
  journal={Nucleic acids research},
  volume={39},
  number={suppl\_1},
  pages={D367--D372},
  year={2010},
  publisher={Oxford University Press}
}

@article{lee2020biobert,
  title={BioBERT: a pre-trained biomedical language representation model for biomedical text mining},
  author={Lee, Jinhyuk and Yoon, Wonjin and Kim, Sungdong and Kim, Donghyeon and Kim, Sunkyu and So, Chan Ho and Kang, Jaewoo},
  journal={Bioinformatics},
  volume={36},
  number={4},
  pages={1234--1240},
  year={2020},
  publisher={Oxford University Press}
}

@article{beltagy2019scibert,
  title={Scibert: A pretrained language model for scientific text},
  author={Beltagy, Iz and Lo, Kyle and Cohan, Arman},
  journal={arXiv preprint arXiv:1903.10676},
  year={2019}
}

@article{gu2021domain,
  title={Domain-specific language model pretraining for biomedical natural language processing},
  author={Gu, Yu and Tinn, Robert and Cheng, Hao and Lucas, Michael and Usuyama, Naoto and Liu, Xiaodong and Naumann, Tristan and Gao, Jianfeng and Poon, Hoifung},
  journal={ACM Transactions on Computing for Healthcare (HEALTH)},
  volume={3},
  number={1},
  pages={1--23},
  year={2021},
  publisher={ACM New York, NY}
}

@article{kingma2014adam,
  title={Adam: A method for stochastic optimization},
  author={Kingma, Diederik P and Ba, Jimmy},
  journal={arXiv preprint arXiv:1412.6980},
  year={2014}
}

@article{shoeybi2019megatron,
  title={Megatron-lm: Training multi-billion parameter language models using model parallelism},
  author={Shoeybi, Mohammad and Patwary, Mostofa and Puri, Raul and LeGresley, Patrick and Casper, Jared and Catanzaro, Bryan},
  journal={arXiv preprint arXiv:1909.08053},
  year={2019}
}

\end{document}